\documentclass[11pt]{article}

\usepackage{microtype}
\usepackage{graphicx}
\usepackage{subcaption}
\usepackage{booktabs}
\usepackage{enumitem}

\usepackage{amsmath}
\usepackage{amssymb}
\usepackage{mathtools}
\usepackage{amsthm}
\usepackage{bm}

\usepackage[numbers,sort&compress]{natbib}

\usepackage[colorlinks=true,linkcolor=blue,citecolor=blue,urlcolor=blue]{hyperref}

\usepackage[capitalize,noabbrev]{cleveref}

\usepackage[margin=1in]{geometry}

\theoremstyle{plain}

\theoremstyle{definition}

\theoremstyle{remark}

\begin{document}

\title{Spectrally Regularized Latent Flow Matching\\for Turbulence Generation}

\author{
  Khalid Rafiq \quad Aditya G. Nair \\[4pt]
  Department of Mechanical Engineering \\
  University of Nevada, Reno, NV, USA \\[4pt]
  \texttt{krafiq@unr.edu}
}

\maketitle

\begin{abstract}
Latent diffusion and flow matching have emerged as leading approaches
for synthetic turbulence generation, yet they systematically
under-represent dissipation-range amplitudes. We introduce a latent
flow matching framework with a spectrally regularized compression
stage that directly targets this failure mode. On a \(256^2\) DNS
dataset at \(Re_f \approx 2250\), replacing an MSE-trained VAE with a
zone-weighted log-spectral objective raises deep-dissipation retained
spectral power from \(25\%\) to \(94\%\) in reconstruction and from
\(20\%\) to \(79\%\) in unconditional generation. The improved latent
representation also yields a substantially better sampling
cost--fidelity tradeoff: the MSE-trained latent space imposes a
fundamental quality ceiling near DD bias \(-0.70\) that no integrator
or step-count can overcome, while the spectrally regularized latent
space reaches DD bias \(-0.117\) at just \(20\) function evaluations.
Mechanistically, encoder--decoder swap experiments show that the
improvement is driven primarily by encoder-induced latent
reorganization rather than decoder capacity, while a
support-amplitude decomposition reveals that MSE-trained models
behave as conservative suppression models, minimizing pointwise error
by attenuating intermittent high-wavenumber structure. Both pipelines
recover the second-order structure function and the correct sign of
\(S_3\), indicating the correct cascade direction without explicit
supervision. A small residual gap in the magnitude of \(S_3\) suggests
that phase-coherent triadic organization remains a complementary axis
to amplitude fidelity for future generative turbulence models.
\end{abstract}

\section{Introduction}
\label{sec:intro}

A practical generator of turbulent flow fields would substantially
reduce the cost of downstream tasks that currently rely on direct
numerical simulation (DNS), including uncertainty quantification,
ensemble statistics, closure-model training, and synthetic inflow
generation. Recent advances have converged on latent generative
pipelines, where diffusion or flow matching models are trained on
learned low-dimensional representations of turbulent states
\citep{rombach2022latent, lipman2023flow, du2024conditional,
liu2025confild, parikh2026conditional, oommen2024diffusion,
whittaker2024turbulence, granero2024multiscale}. Despite their
success, these models exhibit a persistent failure mode: when trained
with pointwise reconstruction objectives, they systematically
under-represent dissipation-range amplitudes. This limitation is
particularly important because high-wavenumber dynamics govern
enstrophy dissipation and strongly influence the evolution of
downstream flow physics \citep{davidson2015turbulence}. More broadly,
the role of the compression objective in shaping downstream generative
dynamics remains poorly understood. In latent generative models, the
encoder does not merely compress the data distribution; it organizes
the geometry of the latent manifold on which sampling and transport
are subsequently performed. We show that modifying this compression
objective can substantially alter both generative fidelity and
sampling efficiency.

We introduce a latent flow matching framework with a spectrally
regularized compression stage based on a zone-weighted log-spectral
objective. The learned latent representation is propagated through a
CondOT-path flow matching generator \citep{lipman2023flow}. On forced
two-dimensional Navier--Stokes turbulence at \(Re_f \approx 2250\) on
a \(256^2\) grid, replacing an MSE-trained variational autoencoder
with the proposed spectral objective improves deep-dissipation
retained spectral power from \(25\%\) to \(94\%\) in reconstruction
and from \(20\%\) to \(79\%\) in unconditional generation. The
improvement also reshapes the latent transport geometry: the
MSE-trained latent space imposes a fundamental quality ceiling near DD
bias \(-0.70\) that no integrator or step-count can overcome, while
the spectrally regularized latent space reaches DD bias \(-0.117\) at
just \(20\) function evaluations.

To understand the source of this gain, we introduce two diagnostics
that probe how compression objectives reshape latent representations.
First, an encoder--decoder swap experiment shows that the improvement
is co-adapted, with encoder-side latent reorganization playing the
dominant role; the baseline decoder is unable to interpret the
reorganized latent code. Second, a support--amplitude decomposition
reveals a structural failure mode of pointwise reconstruction losses:
MSE-trained models act as conservative suppression models, achieving
low pointwise error by systematically attenuating intermittent
dissipation-range structures. In contrast, the spectral objective
restores high-wavenumber amplitudes at a small pointwise cost.
Together, these diagnostics identify where spectral regularization
acts within the architecture and clarify how reconstruction objectives
influence downstream generative behavior.

The same controlled setup also clarifies what the method does not
resolve. Both pipelines recover the second-order structure function
\(S_2(r)\) and the correct \emph{sign} of the third-order structure
function \(S_3(r)\), and hence the direction of the cascade, without
explicit supervision on structure statistics. However, the
\emph{magnitude} of \(S_3(r)\) retains a small residual gap that the
spectral regularizer does not close. This is consistent with the
construction of the objective itself: shell-averaged spectral
penalties constrain Fourier amplitudes but are intrinsically
insensitive to inter-scale phase organization and triadic coherence.
We therefore view phase-coherent interactions as a complementary axis
to amplitude fidelity rather than a competing one.

\paragraph{Contributions.}
We design a controlled two-pipeline study in which all architectural
components are held fixed except the compression objective, enabling
direct isolation of spectral regularization effects. This yields the
following contributions:

\begin{enumerate}[leftmargin=1.2em,topsep=2pt,itemsep=2pt]

\item \textbf{Spectrally consistent generative modeling.}
Spectral regularization substantially improves recovery of fine-scale
structure in unconditional latent generation, raising deep-dissipation
retained spectral power from \(25\%\) to \(94\%\) in reconstruction
and from \(20\%\) to \(79\%\) in generation.

\item \textbf{Improved sampling efficiency through latent geometry.}
The MSE-trained latent space imposes a fundamental quality ceiling
that no integrator or step-count can overcome; the spectrally
regularized latent manifold accesses a strictly better fidelity
regime at as few as \(20\) function evaluations.

\item \textbf{Mechanistic understanding of latent reorganization.}
Encoder--decoder swap experiments demonstrate that the gains arise
primarily from encoder-driven latent reorganization rather than
increased decoder expressivity.

\item \textbf{A failure mode of pointwise reconstruction losses.}
Pointwise objectives favor suppression of intermittent
high-frequency structure, leading to deceptively low reconstruction
error despite poor spectral fidelity.

\item \textbf{Phase coherence as a complementary generative axis.}
Both pipelines recover \(S_2\) and the sign of \(S_3\) without
explicit supervision, while the residual gap in the magnitude of
\(S_3\) identifies phase-coherent triadic organization as a natural
direction for future generative objectives.

\end{enumerate}

\section{Related Work}
\label{sec:related}

\paragraph{Latent generative models for turbulence.}
Deep generative models have been increasingly applied to turbulent
flow fields. Early work focused on GANs
\citep{drygala2022generative, granero2024multiscale}, while more
recent approaches are dominated by diffusion
\citep{ho2020denoising, song2021score} and flow-matching models
\citep{lipman2023flow, liu2023flow, albergo2023stochastic,
shaul2023kinetic}. CoNFiLD \citep{du2024conditional} and
CoNFiLD-inlet \citep{liu2025confild} combine latent diffusion with
neural-field representations of fluid flows, while
\citet{parikh2026conditional} apply conditional flow matching to
wall-bounded turbulence. \citet{whittaker2024turbulence} further show
that diffusion models can recover inertial-range scaling statistics.
Hybrid approaches combining neural operators with diffusion have also
been proposed to improve spectral accuracy
\citep{oommen2024diffusion}.

Across these works, latent compression plays a central role
\citep{nakamura2021convolutional, nakamura2020cnn, doan2021auto,
doan2023convolutional, wiewel2019latent, rafiq2025cluster,
rafiq2025single, kontolati2023ldeepnet}. However, existing pipelines
rely on pointwise (MSE-based) reconstruction objectives, which
systematically under-resolve the dissipation range of the spectrum.
The present work addresses this limitation by introducing a
spectrally regularized compression stage within a latent flow
matching pipeline and showing that the resulting gain propagates from
reconstruction to unconditional generation.

\paragraph{Spectral and physics-aware losses.}
Neural networks exhibit a well-documented bias toward low-frequency
content \citep{rahaman2019spectral}, motivating the use of
frequency-aware objectives. Spectral losses have been explored in
neural-operator forecasting of chaotic systems
\citep{chakraborty2026bsp} and in diffusion models augmented with
operator structure \citep{oommen2024diffusion}.

Our approach differs in two key ways. First, spectral regularization
is applied at the compression bottleneck of a generative model,
rather than as a forecasting penalty. Second, the loss is
zone-weighted in log-spectral space to explicitly target the
disparity between inertial-range and dissipation-range amplitudes.
The encoder--decoder swap and support--amplitude diagnostics
introduced here further provide a mechanistic analysis of how
reconstruction objectives reshape latent geometry and downstream
generative behavior, which has not been examined in prior work.

\paragraph{Subgrid-scale modeling and turbulence super-resolution.}
Two adjacent threads tackle related problems of recovering unresolved
high-wavenumber content. Data-driven subgrid-scale (SGS) closure for
large-eddy simulation seeks to express the effect of unresolved
scales on resolved ones via neural surrogates. Representative
approaches train neural surrogates on filtered DNS to recover
unresolved stresses \citep{maulik2017neural, gamahara2017searching,
beck2019deep, sirignano2020dpm}; later \emph{a-posteriori}-stable
formulations target consistency with the LES rollout itself,
including stability-constrained and physics-consistent
parameterizations for $2$D and quasi-geostrophic turbulence
\citep{park2021toward, guan2022stable, frezat2022posteriori,
kurz2023deep}. Turbulence super-resolution casts scale recovery as
supervised upsampling from a coarse field. Multi-scale CNN
super-resolvers \citep{fukami2019super} have been extended with
physics-informed and adversarial losses \citep{bode2021using,
kim2021unsupervised, yousif2023high}, and recent diffusion- and
score-based methods have been applied to turbulent super-resolution
and reconstruction \citep{shu2023physics, oommen2024diffusion}.

Our setting differs from both threads in two key ways. First, we
study \emph{unconditional} generation rather than coarse-to-fine
reconstruction: dissipation-range structure must emerge from the
latent prior alone. Second, the failure mode we address is structural
rather than algorithmic. The high-\(k\) attenuation observed in
latent generators is induced by the pointwise reconstruction
objective at the compression bottleneck itself, not by filtered-DNS
training or the choice of downsampling operator. The diagnostics
introduced here (spectral-bias zoning, encoder--decoder swap, and
support--amplitude decomposition) generalize naturally to both
settings.

\paragraph{2D turbulence and phase organization.}
The classical theory of two-dimensional turbulence
\citep{kraichnan1967inertial, batchelor1969computation,
kolmogorov1941local, frisch1995turbulence, boffetta2012two,
davidson2015turbulence} describes an inverse energy cascade and a
direct enstrophy cascade mediated by vortex filamentation. This
process is intrinsically phase-coherent: strain induced by one vortex
deforms another into elongated structures. The third-order
longitudinal structure function $S_3(r)$ provides a standard
diagnostic of this inter-scale organization.

\paragraph{Causality and controllable structures.}
Prior work has identified the structures that dominate interaction
and control in 2D turbulence. \citet{jimenez2018machine} and
\citet{jimenez2020monte} show via perturbation experiments that
vortex dipoles, rather than individual vortices, are the most
causally significant structures under velocity-based norms.
Similarly, \citet{yeh2021network} use network-based analysis of
Navier--Stokes interactions to identify dipole regions as dominant
broadcast sites for perturbations and control. These results
emphasize that generators intended for downstream control or
causality use must preserve the joint amplitude--phase organization
of vortex pairs, motivating our use of $S_2$ and $S_3$ as
complementary diagnostics alongside the spectrum.

\section{Problem Setup and Data}
\label{sec:data}

We solve the 2D incompressible Navier--Stokes equations in vorticity
form on the doubly periodic domain $\Omega=[0,2\pi]^2$,
\begin{equation}
\partial_t \omega + (\mathbf{u}\!\cdot\!\nabla)\omega
   = \nu \nabla^2 \omega + f,
\quad \nabla\!\cdot\!\mathbf{u}=0,
\label{eq:ns_vort}
\end{equation}
with $\nu=10^{-3}$ and a body force injecting energy near $k_f=4$.
Simulations are performed using \texttt{jax-cfd}
\citep{kochkov2021machine} on a \(256^2\) grid with fourth-order
time integration and standard \(2/3\) dealiasing, yielding
\(k_{\max}=85\). The dataset is integrated for $T=600$
($1.2\!\times\!10^6$ steps), with snapshots saved every $200$ steps.
The first \(1000\) snapshots are discarded as transient spin-up,
after which the flow reaches a statistically stationary regime. The
remaining $5000$ fields are split temporally into $4500$ train and
$500$ test, and standardized using training statistics.

\begin{figure*}[t]
\centering
\includegraphics[width=0.95\textwidth]{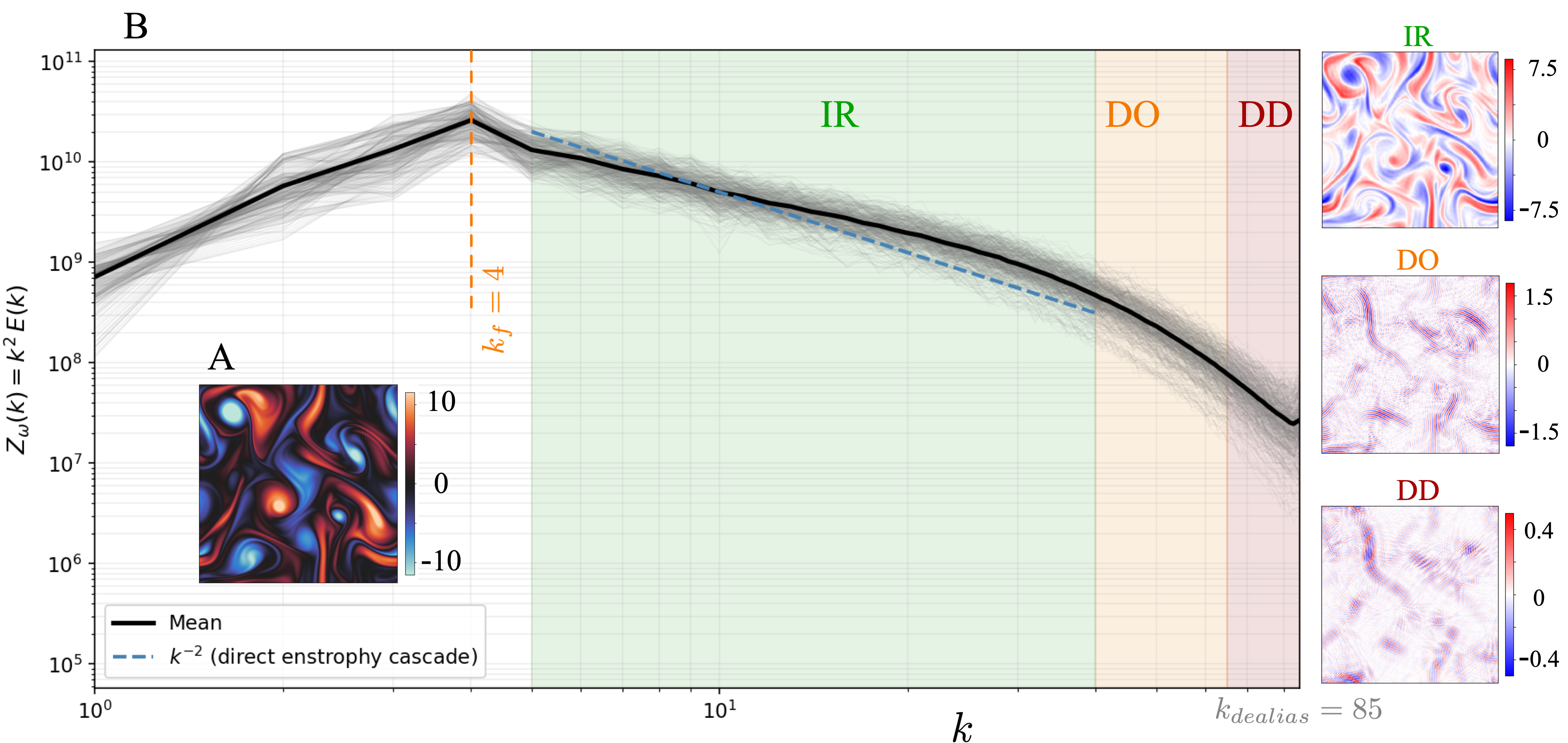}
\caption{Dataset overview and spectral zoning.
\textbf{(A)} Representative standardized vorticity snapshot from the
statistically stationary \(256^2\) DNS at \(Re_f \approx 2250\).
\textbf{(B)} Mean shell-averaged vorticity power
\(Z_\omega(k)=k^2E(k)\) over the test set (black) with individual
realization spectra (grey) and the \(k^{-2}\)
direct-enstrophy-cascade reference slope (blue dashed). The forcing
wavenumber \(k_f=4\) and dealiasing cutoff
\(k_{\mathrm{dealias}}=85\) are marked. Shaded regions indicate the
three spectral zones used throughout the paper: an inertial-range
band IR (\(k=6\text{--}40\), green), a dissipation-onset band DO
(\(k=41\text{--}65\), orange), and a deep-dissipation band DD
(\(k=66\text{--}85\), red). Right panels show representative
band-pass-filtered realizations within each spectral zone. The
characteristic amplitude decreases from \(O(\pm7.5)\) in IR to
\(O(\pm0.4)\) in DD, producing an approximate \(20\times\) signal
imbalance across scales and an effective \(\sim400\times\) disparity
under pointwise \(\ell_2\) reconstruction losses.}
\label{fig:dataset_and_zones}
\end{figure*}

The forcing-scale Reynolds number is
$Re_f = u_{\mathrm{rms}} L_f/\nu \approx 2.25\!\times\!10^3$ with
$u_{\mathrm{rms}}\!\approx\!1.43$, placing the run in a
moderate-Reynolds turbulent regime. The forcing-scale turnover time
is $\tau_f = L_f/u_{\mathrm{rms}}\!\approx\!1.10$, so the retained
dataset spans approximately $456\,\tau_f$. The moderate scale
separation \(k_{\max}/k_f \approx 21\) produces a resolved but
finite cascade range, consistent with prior studies of
moderate-Reynolds-number two-dimensional turbulence
\citep{boffetta2012two}.

\paragraph{Spectral zoning.}
We partition the resolved spectrum into three bands used throughout
the paper, illustrated in \cref{fig:dataset_and_zones}. The
inertial-range band IR (\(k=6\text{--}40\)) contains the dominant
large-scale coherent structures visible in the green-filtered inset
of \cref{fig:dataset_and_zones}, while the dissipation-onset band DO
(\(k=41\text{--}65\)) and the deep-dissipation band DD
(\(k=66\text{--}85\)) progressively isolate finer-scale intermittent
structures. This spectral separation induces a severe imbalance under
pointwise reconstruction objectives. Characteristic IR vorticity
amplitudes are \(O(\pm7.5)\), whereas DD amplitudes are only
\(O(\pm0.4)\), corresponding to an approximate \(20\times\) disparity
in signal magnitude, as shown by the filtered realizations in
\cref{fig:dataset_and_zones}. Under an \(\ell_2\) reconstruction
loss, this becomes an effective \(\sim400\times\) imbalance in
squared-error weighting per spatial location, causing standard
objectives to overwhelmingly prioritize large-scale structure while
systematically underweighting fine-scale content. This imbalance
motivates the zone-weighted spectral regularization introduced in
this work.

\section{Method}
\label{sec:method}

\subsection{Two-stage Pipeline}
\begin{figure}[t]
\centering
\includegraphics[width=0.80\textwidth]{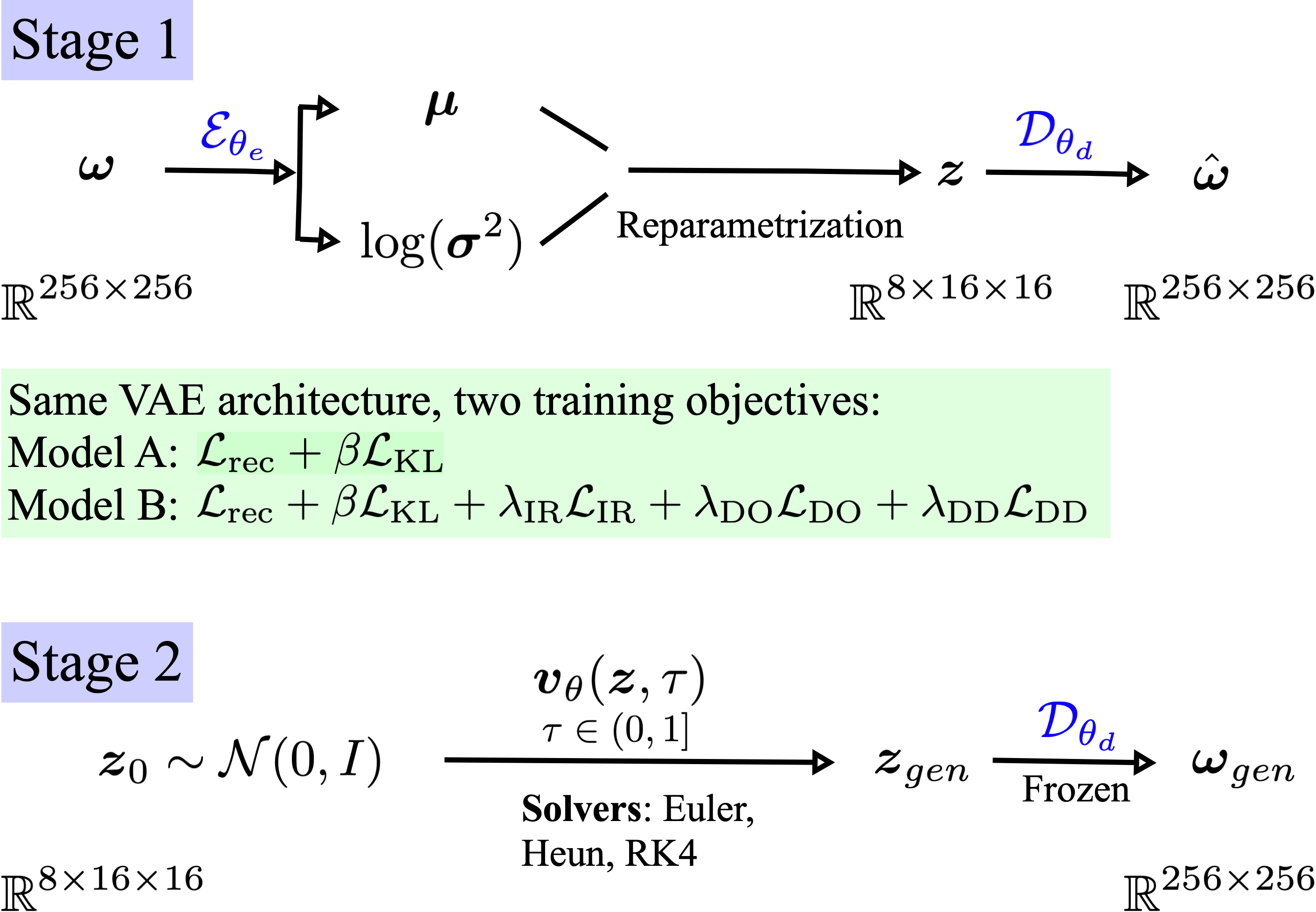}
\caption{Two-stage latent generative pipeline. Stage~1 is a residual
VAE trained with either a baseline objective (Model~A) or a
zone-weighted spectral objective (Model~B). Stage~2 freezes the
decoder and trains a latent CondOT flow matching generator. Sampled
latents are decoded back to vorticity. The two pipelines share
architecture, dataset, and Stage~2 generator exactly, differing only
in compression objective and the resulting latent geometry.}
\label{fig:pipeline_and_zones}
\end{figure}

The framework, summarized in \cref{fig:pipeline_and_zones}, separates
representation learning from latent generative transport. Stage~1 is
a residual VAE that maps a vorticity snapshot
$\omega\!\in\!\mathbb{R}^{1\times256\times256}$ to a structured
latent tensor $z\!\in\!\mathbb{R}^{8\times16\times16}$ (a $32\times$
compression in spatial volume) and reconstructs back. Stage~2 freezes
the decoder, encodes the training set with the encoder mean
$\mu_\phi(\omega)$ \citep{rombach2022latent}, and learns an
unconditional flow-matching model on the resulting latent
representation manifold. Sampling is performed by integrating the
learned vector field forward from a Gaussian prior and decoding the
terminal latent state through the frozen Stage~1 decoder. The key
design hypothesis is that modifying the compression objective
reshapes the geometry of the latent representation on which the flow
matching dynamics are subsequently learned.

We instantiate two compression models with identical architecture and
hyperparameters and differing only in objective. Both encoder and
decoder are multiresolution residual convolutional networks
\citep{he2016deep, wu2018group, hendrycks2016gaussian} with circular
padding to respect periodicity. Both have $\sim\!5$M parameters. The
Stage~2 velocity field $v_\theta(z,\tau)$ is a lightweight
U-Net~\citep{ronneberger2015u} of $\sim\!2$M parameters with
sinusoidal time embedding and self-attention at the bottleneck.
Architecture details appear in \cref{appendix:arch}.

\subsection{Compression Objectives}

\paragraph{Model A (baseline).}
Model~A uses the standard VAE objective:
\begin{equation}
\mathcal{L}_A = \frac{1}{N}\sum_{j}
\left[\,
\|\omega_j-\hat{\omega}_j\|_2^2
+ \beta\,\mathrm{KL}\!\big(q_\phi(z|\omega_j)\,\|\,\mathcal{N}(0,I)\big)
\right].
\label{eq:loss_A}
\end{equation}

\paragraph{Model B (zone-weighted log-spectral).}
Model~B augments \cref{eq:loss_A} with a shell-wise spectral penalty.
For each integer wavenumber shell $\mathcal{S}_k$ define the
shell-averaged vorticity power
\begin{equation}
Z_\omega(k) =
   \frac{1}{|\mathcal{S}_k|}\sum_{(k_x,k_y)\in\mathcal{S}_k}
   |\hat{\omega}(k_x,k_y)|^2.
\end{equation}
For zone
$\mathcal{K}_z\in\{\mathrm{IR},\mathrm{DO},\mathrm{DD}\}$ the zone
loss is the mean squared log-spectral error,
\begin{equation}
\mathcal{L}_z =
   \frac{1}{|\mathcal{K}_z|}\sum_{k\in\mathcal{K}_z}
   \big[\log(Z_{\hat{\omega}}(k)+\epsilon)
        - \log(Z_\omega(k)+\epsilon)\big]^2,
\label{eq:zone_loss}
\end{equation}
giving the full objective
\begin{equation}
\mathcal{L}_B = \mathcal{L}_A
   + \lambda_{\mathrm{IR}}\mathcal{L}_{\mathrm{IR}}
   + \lambda_{\mathrm{DO}}\mathcal{L}_{\mathrm{DO}}
   + \lambda_{\mathrm{DD}}\mathcal{L}_{\mathrm{DD}}.
\label{eq:loss_B}
\end{equation}

\cref{eq:zone_loss} constrains the modulus of the Fourier transform
on shells of constant $|\mathbf{k}|$. It is invariant to (a) the
phase of any individual mode, (b) the relative phase between modes,
and (c) the distribution of energy among modes within a shell. It
cannot enforce inter-scale phase organization or triadic coherence.

\subsection{Latent Flow Matching}
\label{sec:fm}
We use the linear conditional optimal transport (CondOT) probability
path \citep{lipman2023flow}. Given $z_1\!\sim\!q(z)$,
$\varepsilon\!\sim\!\mathcal{N}(0,I)$,
$\tau\!\sim\!\mathcal{U}[0,1]$, the interpolant
$z_\tau = (1-\tau)\varepsilon + \tau z_1$ has target velocity
$u_\tau(z_\tau|z_1) = z_1 - \varepsilon$, and the loss is
\begin{equation}
\mathcal{L}_{\mathrm{FM}}(\theta) =
   \mathbb{E}_{\tau,z_1,\varepsilon}\big\|
   v_\theta(z_\tau,\tau) - (z_1-\varepsilon)\big\|_2^2.
\label{eq:fm_loss}
\end{equation}

At inference, $z_0\!\sim\!\mathcal{N}(0,T^2 I)$ is integrated
forward by
\begin{equation}
\frac{d z_\tau}{d\tau} = v_\theta(z_\tau,\tau),
\quad \tau\in[0,1],
\label{eq:fm_ode}
\end{equation}
and the terminal state is decoded. The scalar $T$ corrects a mild
empirical under-dispersion fitted from 500 draws from the Gaussian
prior $\mathcal{N}(0, I)$. The nearly identical values
$T_A\!=\!1.157$ and $T_B\!=\!1.170$ confirm the calibration does not
explain the gap between pipelines. CondOT is chosen because its
low-kinetic-energy trajectories enable accurate generation at low NFE
\citep{shaul2023kinetic}.

\paragraph{Training.}
Both VAEs are trained for \(200\) epochs with batch size \(48\),
learning rate \(7.5\times10^{-4}\), and \(\beta=7.5\times10^{-3}\).
For Model~B, spectral weights are selected via Bayesian sweep,
yielding the ratio \(1:4:6\) for
\((\lambda_{\mathrm{IR}},\lambda_{\mathrm{DO}},\lambda_{\mathrm{DD}})\).
A uniform spectral weighting
\((\lambda_{\mathrm{IR}}{=}\lambda_{\mathrm{DO}}{=}
\lambda_{\mathrm{DD}}{=}4\times10^{-3})\) was also evaluated and
improved over the MSE-only baseline, but the zone-weighted
\((1\!:\!4\!:\!6)\) configuration achieves lower bias across all
three spectral zones. Stage~2 flow-matching models are trained for
\(1000\) epochs with batch size \(64\) and learning rate
\(3\times10^{-4}\).

\section{Stage 1: Effects of Spectral Regularization}
\label{sec:stage1}

\subsection{Ensemble Spectral Fidelity}

\begin{figure}[t]
\centering
\includegraphics[width=0.90\textwidth]{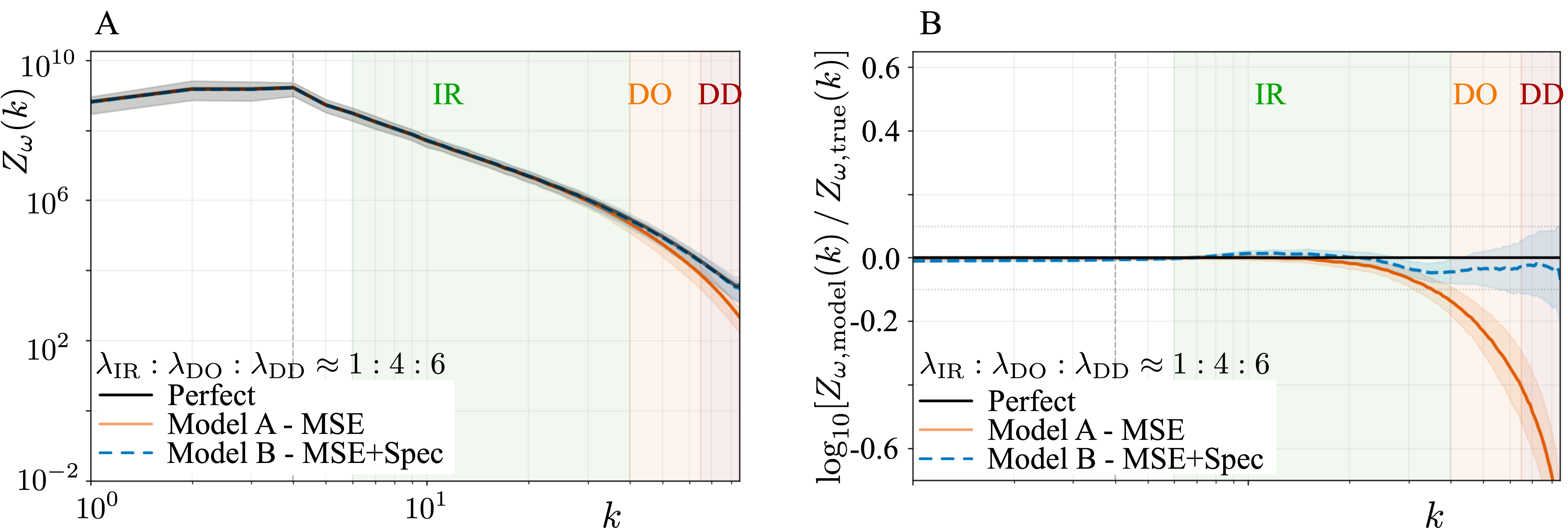}
\caption{Stage~1 ensemble spectral fidelity.
\textbf{Left:} mean shell-averaged $Z_\omega(k)$ over the held-out
test set with 10--90\% bands.
\textbf{Right:} spectral bias mean $\pm 1\sigma$. Spectral
regularization in Model~B moves the bias closer to zero in all three
zones, with the largest gain in DD. Both models recover the IR slope.}
\label{fig:vae_spectra}
\end{figure}

\cref{fig:vae_spectra} reports the held-out shell-averaged vorticity
spectrum and per-wavenumber bias,
\begin{equation*}
\log_{10}\!\left[Z_{\omega,\mathrm{model}}(k)\,/\,Z_{\omega,\mathrm{true}}(k)\right].
\end{equation*}

Model~A exhibits a small negative bias in IR and a strong high-$k$
deficit; Model~B is close to zero across all three zones with
narrower uncertainty. Quantitatively (\cref{tab:spectral_bias}),
Model~A preserves only $54\%$ of true DO and $25\%$ of true DD power,
while Model~B preserves $92\%$ and $93\%$ respectively.

\begin{table}[t]
\centering
\small
\setlength{\tabcolsep}{5pt}
\renewcommand{\arraystretch}{1.1}
\caption{Zone-averaged spectral statistics for the two compression
models on the held-out test set. ``ret.'' is the retained spectral
power $100\!\times\!10^{\mathrm{bias}}$ (closer to $100\%$ is
better).}
\label{tab:spectral_bias}
\begin{tabular}{lccccccccc}
\toprule
& \multicolumn{3}{c}{IR} & \multicolumn{3}{c}{DO}
& \multicolumn{3}{c}{DD}\\
\cmidrule(lr){2-4}\cmidrule(lr){5-7}\cmidrule(lr){8-10}
& bias & spr. & ret. & bias & spr. & ret.
& bias & spr. & ret.\\
\midrule
A & $-0.04$ & $0.05$ & $90.8$
  & $-0.27$ & $0.11$ & $54.1$
  & $-0.61$ & $0.19$ & $24.8$\\
B & $\bm{-0.01}$ & $\bm{0.03}$ & $\bm{97.1}$
  & $\bm{-0.03}$ & $\bm{0.06}$ & $\bm{92.3}$
  & $\bm{-0.03}$ & $\bm{0.11}$ & $\bm{93.6}$\\
\bottomrule
\end{tabular}
\end{table}

\subsection{Suppression vs.\ Recovery in the Dissipation Range}
\label{sec:dd_decomp}
\begin{figure}[t]
\centering
\includegraphics[width=0.90\textwidth]{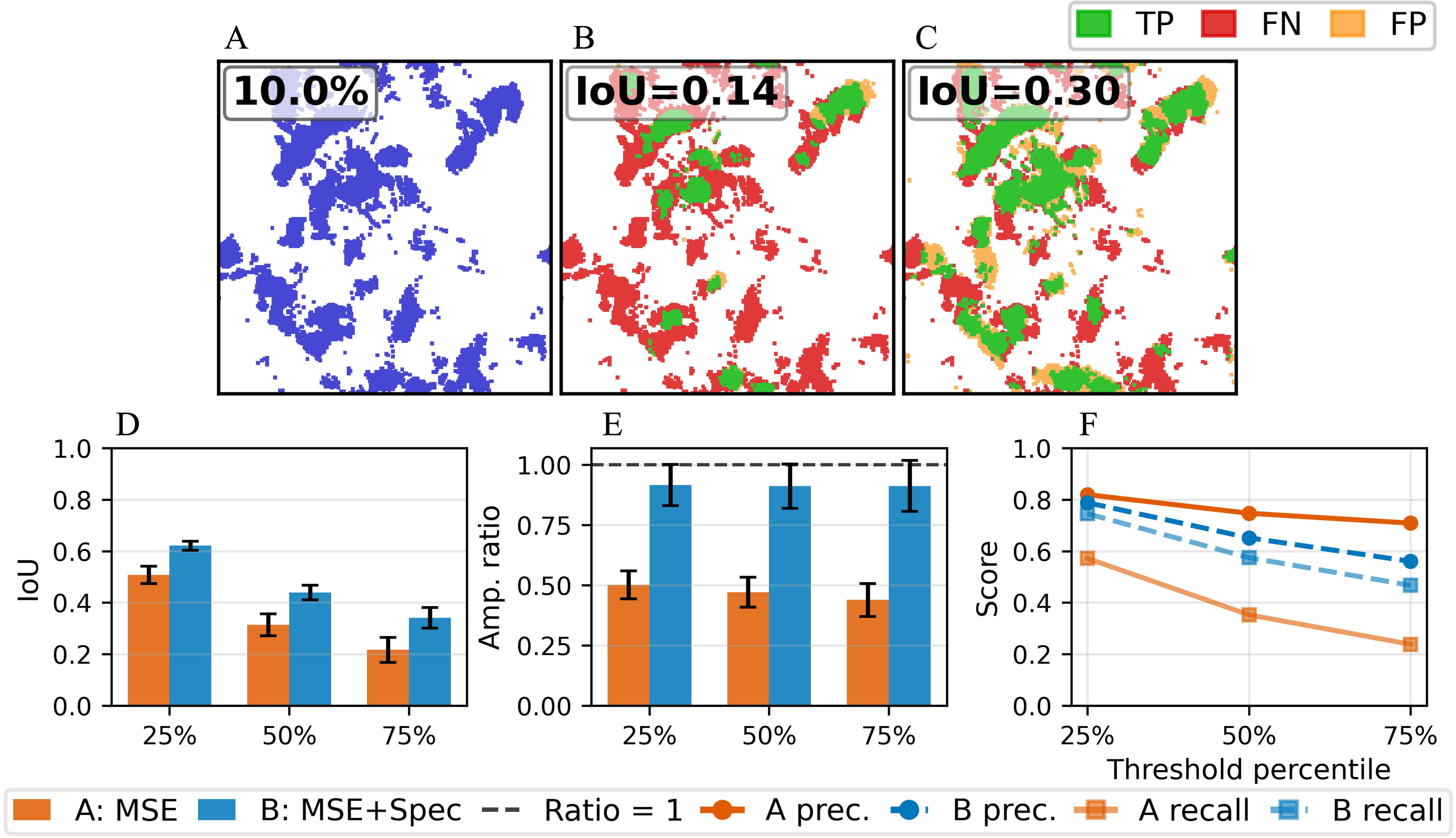}
\caption{Support--amplitude decomposition in the DD band
($k\!=\!66\text{--}85$). \textbf{(A)} Thresholded DD-support mask
from one true field. \textbf{(B--C)} Predictions of Models A and B
as TP (green), FN (red), FP (orange). \textbf{(D)} Ensemble IoU
across thresholds. \textbf{(E)} Amplitude ratio (model/truth) on the
union support. \textbf{(F)} Precision and recall. Model~A is precise
but strongly suppressed (amplitude ratio $\approx 0.44$); Model~B
has higher IoU, much higher recall, and an amplitude ratio $\approx
0.91$ at the cost of broader support.}
\label{fig:dd_decomp}
\end{figure}

Better Fourier-space fidelity does not imply lower pointwise error in
the DD band. Across the test ensemble Model~B has a slightly but
systematically \emph{larger} DD-band MSE than Model~A
($\mu\!=\!6.7\!\times\!10^{-3}$ vs.\
$6.2\!\times\!10^{-3}$); see \cref{appendix:stage1}. To explain this
dissociation, we threshold the band-pass DD field at the $p$-th
percentile of true DD magnitude, producing binary support masks, and
decompose model predictions into true positives, false negatives, and
false positives on the union of active support (\cref{fig:dd_decomp}).

The two pipelines behave qualitatively differently. Model~A acts as a
\emph{conservative-suppression} model: predicting near-zero in sparse
DD regions minimizes MSE with little penalty, systematically
suppressing true support and amplitudes by $\sim\!2\times$. Model~B
behaves as a \emph{recovery} model, restoring most of the true
support and amplitude budget at the cost of a slightly larger
pointwise error. Low MSE on intermittent fine-scale structure can
therefore reflect suppression rather than faithful recovery.

\subsection{Where Does the Spectral Gain Live?}
\begin{figure}[t]
\centering
\includegraphics[width=0.80\textwidth]{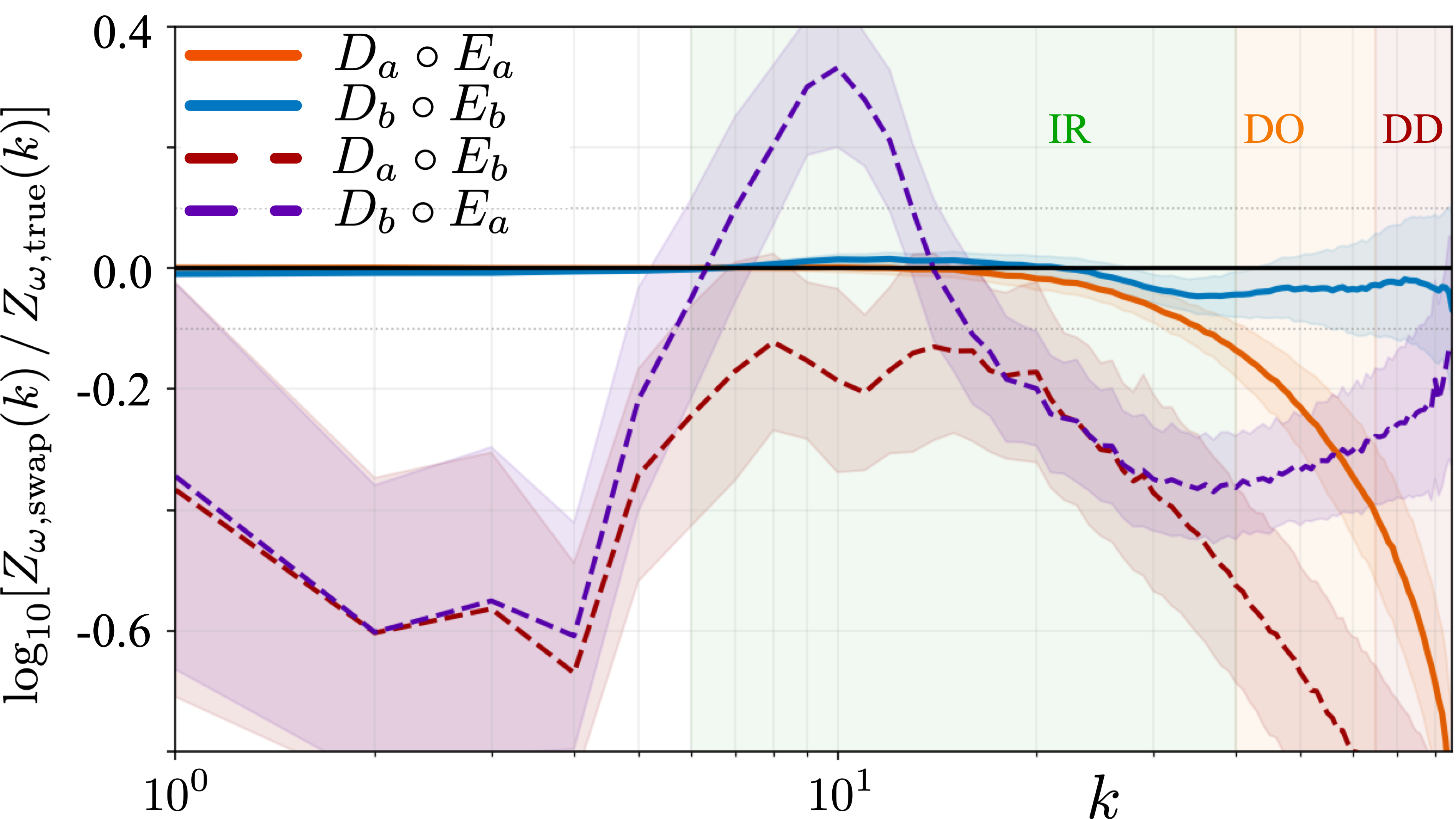}
\caption{Encoder--decoder swap diagnostic. Spectral bias for the
four combinations of
$\{\mathcal{E}_A,\mathcal{E}_B\}\!\times\!\{\mathcal{D}_A,\mathcal{D}_B\}$.
Only the matched pair $\mathcal{D}_B\!\circ\!\mathcal{E}_B$ holds
low bias across all zones. The mismatch
$\mathcal{D}_A\!\circ\!\mathcal{E}_B$ catastrophically degrades,
showing that the baseline decoder cannot reliably interpret the
spectrally regularized latent representation.}
\label{fig:swap}
\end{figure}

To localize the spectral benefit, we evaluate all four pairings of
$\{\mathcal{E}_A,\mathcal{E}_B\}\!\times\!\{\mathcal{D}_A,\mathcal{D}_B\}$
(\cref{fig:swap}, \cref{tab:swap}). The matched spectrally regularized
pair $\mathcal{D}_B\!\circ\!\mathcal{E}_B$ is the only configuration
that holds low bias across all three zones. Cross-swapped pairs are
catastrophic or partial:
$\mathcal{D}_A\!\circ\!\mathcal{E}_B$ is worse than the baseline in
every zone (DD bias $-0.96$), indicating that
$\mathcal{E}_B$'s latent representation is reorganized into a form
$\mathcal{D}_A$ cannot decode. The reverse swap
$\mathcal{D}_B\!\circ\!\mathcal{E}_A$ partially recovers DD ($-0.23$
vs.\ $-0.61$) but is worse than baseline in IR and DO. The spectral
benefit is therefore co-adapted, with an asymmetric encoder-side
anchor: encoder-side latent reorganization is the more fundamental
component, and the decoder contributes complementary but limited
recovery capacity.

\begin{table}[t]
\centering
\small
\setlength{\tabcolsep}{6pt}
\renewcommand{\arraystretch}{1.1}
\caption{Zone-averaged spectral bias for all four encoder--decoder
pairings.}
\label{tab:swap}
\begin{tabular}{lcccccc}
\toprule
& \multicolumn{2}{c}{IR} & \multicolumn{2}{c}{DO}
& \multicolumn{2}{c}{DD}\\
\cmidrule(lr){2-3}\cmidrule(lr){4-5}\cmidrule(lr){6-7}
& bias & spr. & bias & spr. & bias & spr.\\
\midrule
$\mathcal{D}_A\!\circ\!\mathcal{E}_A$
  & $-0.042$ & $0.049$ & $-0.267$ & $0.114$ & $-0.606$ & $0.192$\\
$\mathcal{D}_B\!\circ\!\mathcal{E}_B$
  & $\bm{-0.013}$ & $\bm{0.031}$ & $\bm{-0.035}$ & $\bm{0.062}$
  & $\bm{-0.029}$ & $\bm{0.112}$\\
$\mathcal{D}_A\!\circ\!\mathcal{E}_B$
  & $-0.286$ & $0.198$ & $-0.702$ & $0.203$ & $-0.961$ & $0.196$\\
$\mathcal{D}_B\!\circ\!\mathcal{E}_A$
  & $-0.171$ & $0.243$ & $-0.321$ & $0.098$ & $-0.228$ & $0.150$\\
\bottomrule
\end{tabular}
\end{table}

\section{Stage 2: Spectrally Faithful Generation at Low Cost}
\label{sec:stage2}

\subsection{Generated Spectra Inherit the Stage 1 Gain}
\begin{figure}[t]
\centering
\includegraphics[width=0.90\textwidth]{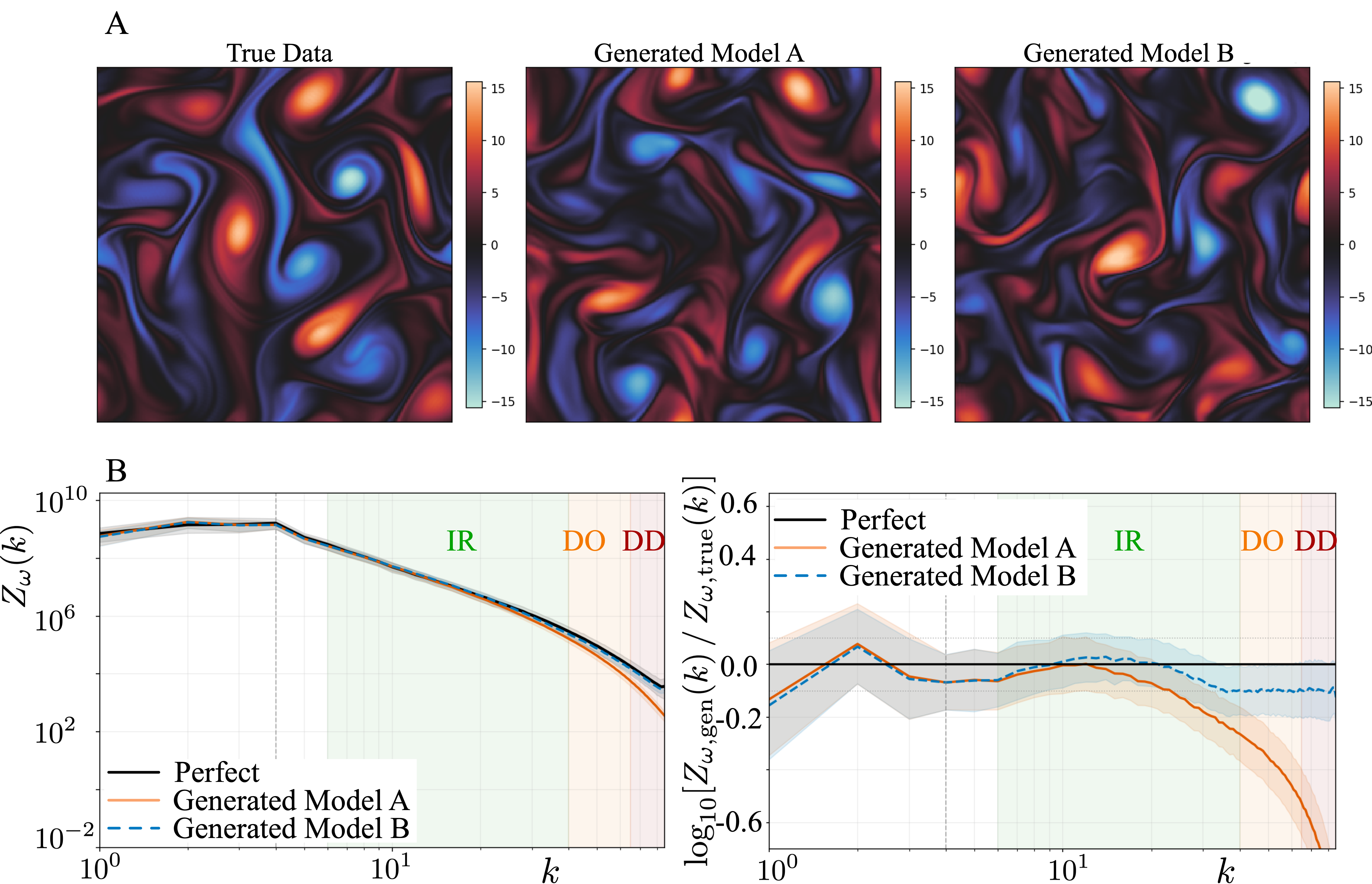}
\caption{Stage~2 generation. \textbf{Top:} representative held-out
test field and unconditional samples from each pipeline. The samples
are draws from the learned latent distribution, not reconstructions.
\textbf{Bottom:} mean shell-averaged spectrum and bias over $500$
generated samples vs.\ $500$ test fields.}
\label{fig:gen}
\end{figure}

\cref{fig:gen} (top) shows representative unconditional samples from
both pipelines together with ensemble shell-averaged spectra over
$500$ generated and $500$ test fields. Both pipelines produce
visually plausible fields with realistic large-scale organization.
The spectra echo the Stage~1 result: Model~A holds the dominant
scales but develops a strong high-$k$ deficit; Model~B remains close
to truth across all three zones (IR bias $-0.03$ vs.\ $-0.10$, DO
$-0.10$ vs.\ $-0.36$, DD $-0.10$ vs.\ $-0.70$). The spectral
regularization of the encoder propagates through to the generative
distribution, raising DD retained spectral power from $20\%$ to
$79\%$ in generation.

\begin{table}[t]
\centering
\small
\setlength{\tabcolsep}{5pt}
\renewcommand{\arraystretch}{1.1}
\caption{Stage~2 generative spectral bias and retained power over
$500$ generated vs.\ $500$ test fields.}
\label{tab:gen}
\begin{tabular}{lccccccccc}
\toprule
& \multicolumn{3}{c}{IR} & \multicolumn{3}{c}{DO}
& \multicolumn{3}{c}{DD}\\
\cmidrule(lr){2-4}\cmidrule(lr){5-7}\cmidrule(lr){8-10}
& bias & spr. & ret. & bias & spr. & ret.
& bias & spr. & ret.\\
\midrule
A & $-0.10$ & $0.13$ & $79.8$
  & $-0.36$ & $0.14$ & $43.8$
  & $-0.70$ & $0.20$ & $20.0$\\
B & $\bm{-0.03}$ & $\bm{0.10}$ & $\bm{92.5}$
  & $\bm{-0.10}$ & $\bm{0.10}$ & $\bm{79.6}$
  & $\bm{-0.10}$ & $\bm{0.11}$ & $\bm{79.4}$\\
\bottomrule
\end{tabular}
\end{table}

\subsection{Sampling Cost--Fidelity and Latent Quality Ceiling}
\label{sec:solvers}

\begin{figure}[t]
\centering
\includegraphics[width=0.90\textwidth]{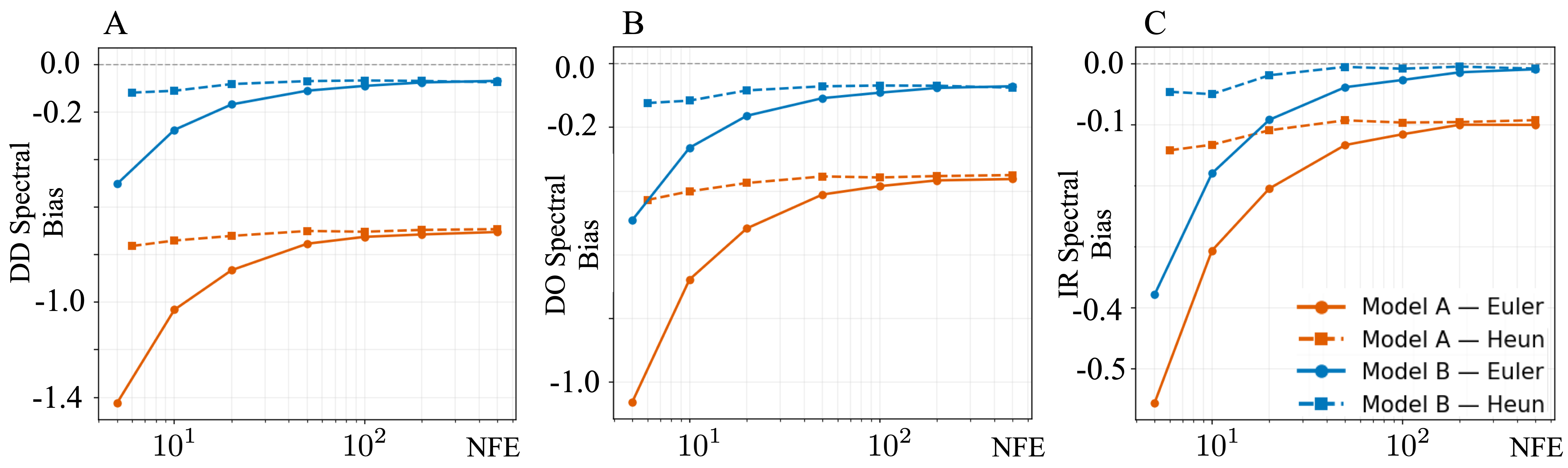}
\caption{Sampling cost--fidelity tradeoff. Spectral bias vs.\ NFE in
the DD (left), DO (middle), and IR (right) zones. Solid: Euler.
Dashed: Heun. Model~A saturates near DD bias $-0.70$ regardless of
integrator or NFE, revealing a fundamental quality ceiling of the
MSE-trained latent space. Model~B Heun reaches DD bias $-0.117$ at
NFE${=}20$; a fidelity level entirely inaccessible to Model~A at any
sampling budget.}
\label{fig:solvers}
\end{figure}

\cref{fig:solvers} reports the Stage~2 spectral bias as a function
of the number of function evaluations (NFE) for fixed-step Euler and
Heun integrators, by zone. On an NVIDIA RTX 3090 with batch $50$,
generation costs $\sim\!3.4$\,ms per NFE. Across all NFE budgets and
all three zones, Model~B has a strictly more favourable
cost--fidelity curve than Model~A. Critically, Model~A Heun
saturates near DD bias $-0.70$ from NFE${=}20$ onwards, matching
Model~A Euler at convergence; indicating that the MSE-trained latent
space imposes a fundamental quality ceiling that no integrator or
step-count can overcome. Model~B Heun reaches DD bias $-0.117$ at
NFE${=}20$ and holds it, accessing a fidelity regime entirely
unavailable to Model~A. Heun reaches this converged accuracy at a
small fraction of the Euler cost; RK4 produces results
indistinguishable from Heun at equal NFE and is omitted, consistent
with the smooth low-kinetic trajectories of the CondOT
path~\citep{shaul2023kinetic}. Model~B with Heun at NFE${=}20$ is
the practical sweet spot.

The result is particularly striking because the Stage~2 generators
are architecturally identical; only the loss function changes which
affects the latent geometry. The ceiling reflects a representation
bottleneck rather than an optimization failure: Model~A's Stage~1
reconstructions exhibit the same DD deficit
(\cref{tab:spectral_bias}), confirming that fine-scale information is
absent from the latent code and cannot be recovered by any downstream
sampler.

\subsection{Structure Functions: Amplitude and Phase}
\label{sec:phase}
The same controlled setup that delivers the amplitude and
cost--fidelity gains lets us probe the phase content of the generated
fields, which the shell-averaged regularizer does not constrain by
construction. We compute longitudinal velocity-increment structure
functions $S_p(r)\!=\!\langle[\delta u_L(\mathbf{x},r)]^p\rangle$
for $p\!=\!2,3$, recovering velocity from generated vorticity via
$\hat{\psi}=-\hat{\omega}/k^2$ and averaging over $24$ azimuthal
directions and $500$ realizations. Here $S_2(r)$ and $S_3(r)$ are
the second- and third-order longitudinal velocity-increment structure
functions; $S_3$ is negative in 2D turbulence and its sign encodes
the direction of the enstrophy cascade. Two cautions apply. First,
dividing by $k^2$ in Fourier space damps high-$k$ content in
velocity, so velocity-increment statistics test a different aspect of
the field than vorticity-shell spectra. Second, $S_3$ is sensitive to
phase-coherent triadic interactions
\citep{frisch1995turbulence, davidson2015turbulence}, the property a
shell-averaged loss cannot enforce.

\begin{figure}[t]
\centering
\includegraphics[width=0.90\textwidth]{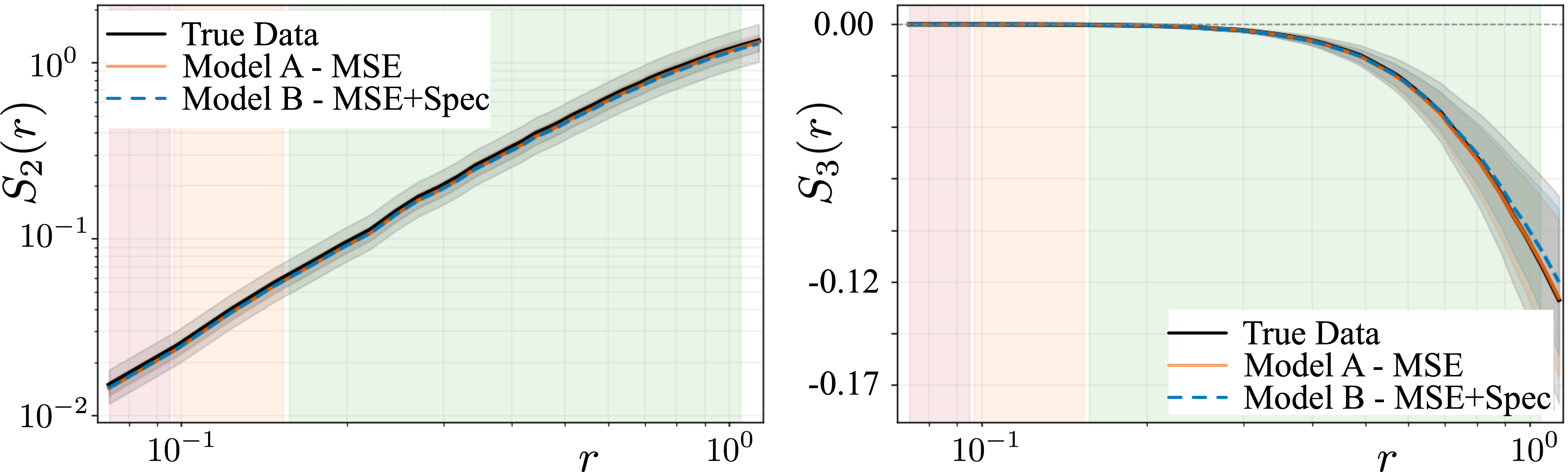}
\caption{Longitudinal structure functions over $500$ realizations.
\textbf{Left:} $S_2(r)$. Both pipelines reproduce the second-order
structure function across the resolved range. \textbf{Right:}
$S_3(r)$. Both pipelines correctly recover the negative sign of
$S_3$ and therefore the direction of the cascade without any
structure-function supervision, with magnitudes close to the test
ensemble. The shell-averaged regularizer is not designed to enforce
phase coherence, so the small residual gap is an expected
complementary direction rather than a failure of the present method.}
\label{fig:s23}
\end{figure}

\cref{fig:s23} reports the result. Both pipelines reproduce $S_2(r)$
across the resolved range and recover the correct \emph{sign} of
$S_3(r)$ throughout the inertial range, i.e., the direction of the
cascade, without any explicit supervision on structure functions.
This indicates that the learned latent dynamics preserve sufficient
multiscale organization for the correct cascade direction to emerge
without explicit supervision. The magnitudes of $S_2$ and $S_3$ are
also in close agreement with the test ensemble, with a small residual
gap in the deepest part of $S_3$.

\paragraph{Reading the small gap.}
The construction of \cref{eq:zone_loss} is informative here: a
shell-averaged loss controls $|\hat{\omega}(\mathbf{k})|^2$ on
annuli of constant $|\mathbf{k}|$ and is invariant to the relative
phase between modes within a shell and across shells. We therefore
expect the spectral regularizer to buy amplitude fidelity but not,
by itself, to enforce inter-scale phase coherence. The fact that
$S_3$ is nonetheless captured in sign and approximately in magnitude
indicates that phase organization is largely recovered through the
latent dynamics, while the residual gap is consistent with
phase-coherent triadic interactions being only implicitly recovered
through the MSE component of the objective
\citep{chakraborty2026bsp}. Closing this remaining gap likely
requires objectives that explicitly constrain phase-coherent
multiscale interactions.

\section{Discussion}
\label{sec:discussion}

\paragraph{Latent geometry and transport conditioning.}
The encoder--decoder swap experiments point to a broader observation
about latent generative models for physical systems. The encoder does
not merely compress the data distribution; it organizes the transport
geometry on which the downstream generative dynamics are learned. Our
results show that modifying the reconstruction objective reorganizes
this latent geometry into a form that is substantially better
conditioned for ODE-based transport and low-NFE sampling. The
co-adaptation observed in the swap experiments further suggests that
phase-sensitive refinement is most naturally introduced as a joint
representation-learning objective, rather than as a correction
applied only to the decoder or generative dynamics. Higher-order
structure-function penalties, two-point correlation objectives, or
bispectral constraints \citep{frisch1995turbulence} are natural
extensions in this direction. More broadly, the results suggest that
reconstruction objectives can substantially alter the conditioning
of latent transport problems in generative models of multiscale
physical systems. In the present setting, spectral regularization
improves not only reconstruction fidelity but also the geometry on
which latent transport trajectories are learned, leading to
substantially improved low-NFE sampling behavior despite identical
Stage~2 architectures.

\paragraph{Implications for downstream physical modeling.}
Generated turbulent ensembles are increasingly used for uncertainty
quantification, surrogate training, synthetic inflow generation, flow
control, and causality analysis. These downstream tasks depend not
only on spectral amplitudes but also on the joint phase organization
of coherent structures such as vortex dipoles. Our results show that
spectrally regularized latent generation substantially improves
fine-scale amplitude fidelity while preserving the correct cascade
direction and large-scale structure-function behavior. The remaining
gap in the magnitude of \(S_3\) suggests that accurately reproducing
phase-coherent multiscale interactions remains an important direction
for future physics-aware generative objectives.

This distinction is particularly relevant for downstream analyses
based on flow interactions rather than pointwise statistics. Prior
work on broadcast-mode flow control \citep{yeh2021network} and
causality analysis \citep{jimenez2018machine, jimenez2020monte} has
shown that dynamically significant structures emerge from the
collective organization of vortex pairs and interaction networks.
Generative models intended for these applications must therefore
recover not only spectral amplitudes but also the relative phase
organization underlying coherent multiscale dynamics.

\paragraph{Limitations and future work.}
The present study uses a single Reynolds number
(\(Re_f \approx 2250\)) on a \(256^2\) domain. Extending the
analysis to higher Reynolds numbers, where the inertial range
broadens and triadic interactions span larger spectral ranges, is a
natural next step. The current implementation also uses a simple
post-hoc variance calibration; more principled latent priors, latent
whitening strategies, and adaptive transport paths remain promising
directions for future work. The analysis is restricted to
two-dimensional turbulence, where cascade structure and coherent
interactions are comparatively well understood; extending to
three-dimensional flows introduces vortex stretching and modified
cascade dynamics, but does not alter the central observation:
pointwise reconstruction objectives systematically bias the geometry
of learned latent representations and underweight fine-scale
multiscale structure; a bias that can be directly remedied by
upweighting small-scale content in the spectral domain.

\section{Conclusion}
\label{sec:conclusion}

We introduced a spectrally regularized latent flow matching framework
for generative modeling of multiscale physical fields and showed that
modifying the compression objective substantially alters downstream
generative behavior. A zone-weighted log-spectral regularizer at the
latent bottleneck yields $93\%$ retention of true DD spectral power
in reconstruction and $79\%$ in generation, while the MSE-trained
latent space imposes a fundamental quality ceiling that no integrator
or step-count can overcome.

Beyond improved spectral fidelity, the results reveal a mechanistic
link between reconstruction objectives, latent geometry, and
transport dynamics. Encoder--decoder swap experiments show that the
gains arise primarily from encoder-driven latent reorganization,
while the support--amplitude analysis demonstrates that pointwise
reconstruction losses can achieve low error through systematic
suppression of intermittent fine-scale structure. The spectrally
regularized latent manifold is also substantially better conditioned
for low-NFE ODE-based sampling despite identical Stage~2
architectures.

Both pipelines recover \(S_2(r)\) and the correct sign of \(S_3(r)\)
without explicit supervision, while the remaining gap in \(S_3\)
magnitude highlights phase-coherent multiscale interactions as a
natural target for future phase-aware objectives. More broadly, the
results suggest that reconstruction objectives can fundamentally
reshape latent transport geometry, downstream sampling dynamics, and
physical fidelity in generative models of multiscale systems.

\section*{Acknowledgements}
A.G.N.\ acknowledges support from the National Science Foundation
AI Institute in Dynamic Systems (Award No.\ 2112085, Program
Manager: Dr.\ Shahab Shojaei-Zadeh). The authors gratefully
acknowledge Professor Ping Liu from the Department of Computer
Science and Engineering at the University of Nevada, Reno,
for helpful and critical discussions.

\bibliographystyle{abbrvnat}
\bibliography{references}

\appendix

\section{Architecture Details}
\label{appendix:arch}
\begin{figure*}[h]
\centering
\includegraphics[width=0.90\textwidth]{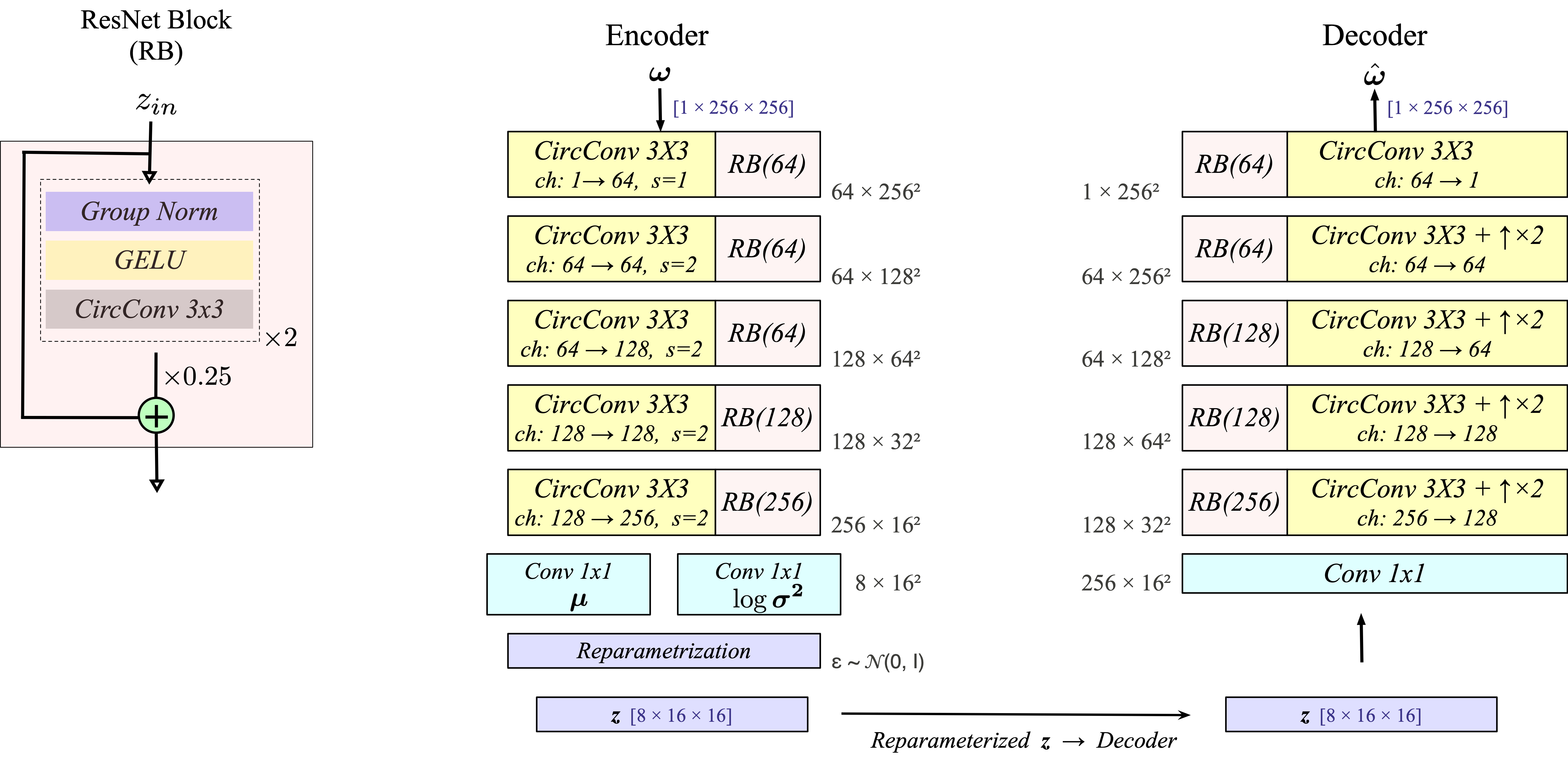}
\caption{Residual VAE architecture used in Stage~1. Models~A and~B
share the architecture exactly and differ only in the training
objective.}
\label{fig:vae_arch}
\end{figure*}

\begin{figure*}[h]
\centering
\includegraphics[width=0.90\textwidth]{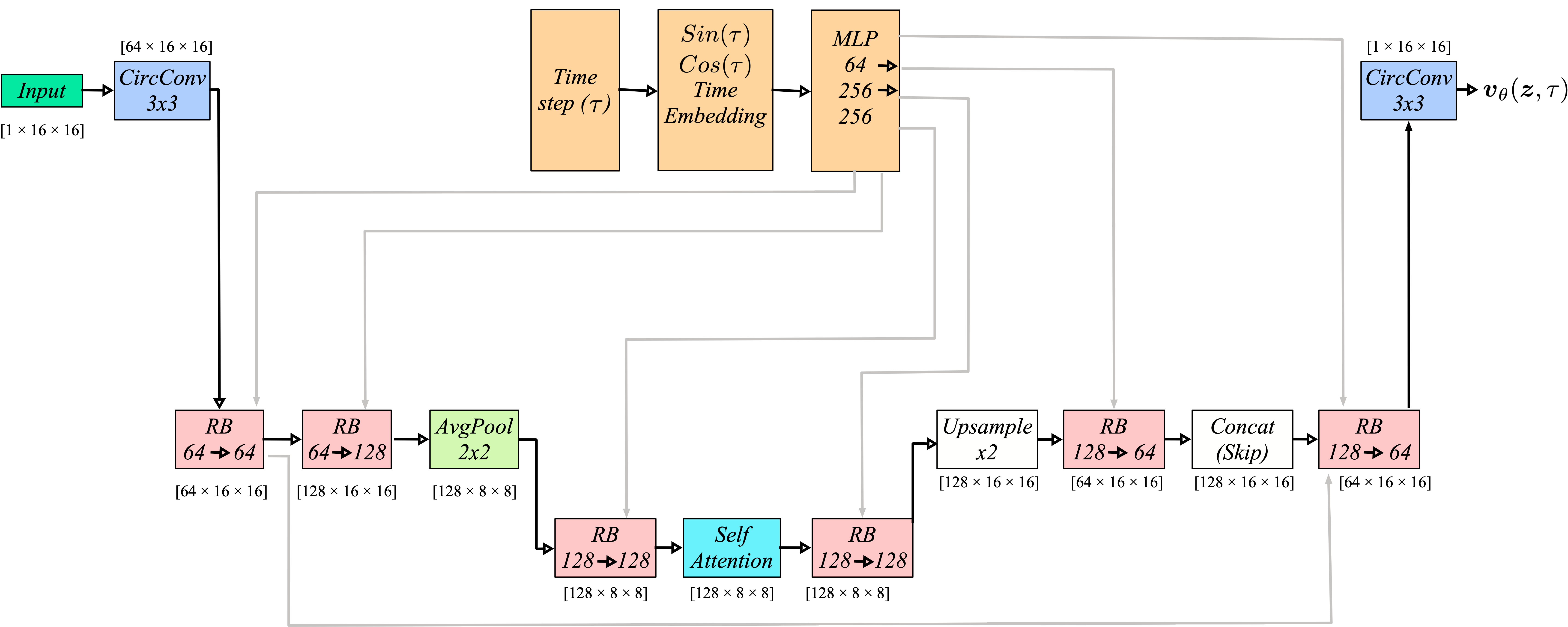}
\caption{U-Net parameterization of the latent flow matching velocity
field $v_\theta(z,\tau)$.}
\label{fig:fm_unet}
\end{figure*}

The Stage~1 VAE architecture is shown in \cref{fig:vae_arch}. The
Stage~2 latent flow matching U-Net is shown in \cref{fig:fm_unet}.
A larger latent of size $8\times32\times32$ (a weaker $16\times$
compression) was also tested; several latent channels exhibited
near-zero empirical variance after training, indicating
representational redundancy without performance benefit.

\section{Additional Stage 1 Diagnostics}
\label{appendix:stage1}
\begin{figure*}[h]
\centering
\includegraphics[width=0.90\textwidth]{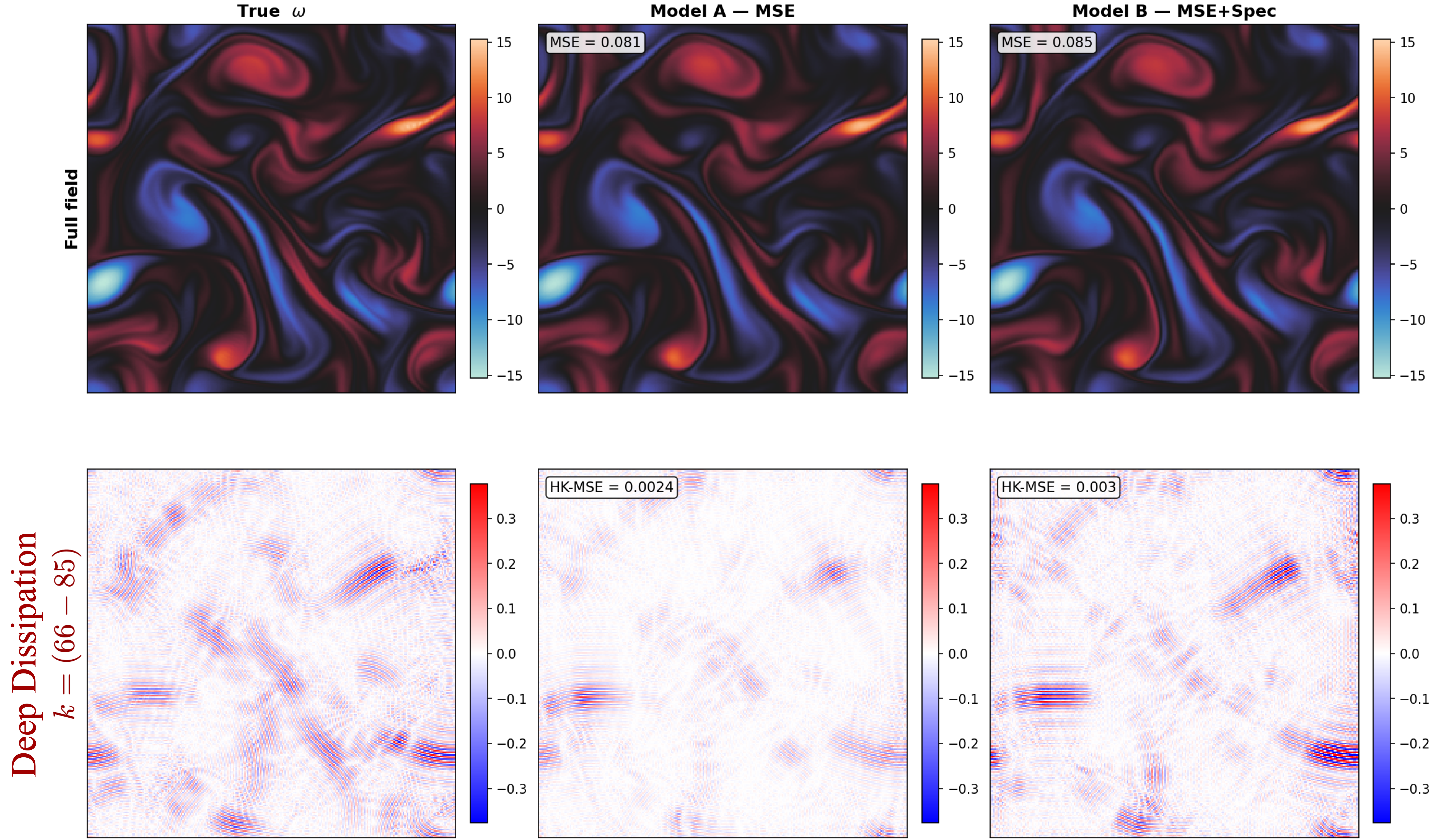}
\caption{Qualitative reconstruction comparison. Top: full field.
Bottom: DD-band ($k\!=\!66\text{--}85$) with color scale locked to
the true field.}
\label{fig:vae_qualitative}
\end{figure*}

\begin{figure*}[h]
\centering
\includegraphics[width=0.90\textwidth]{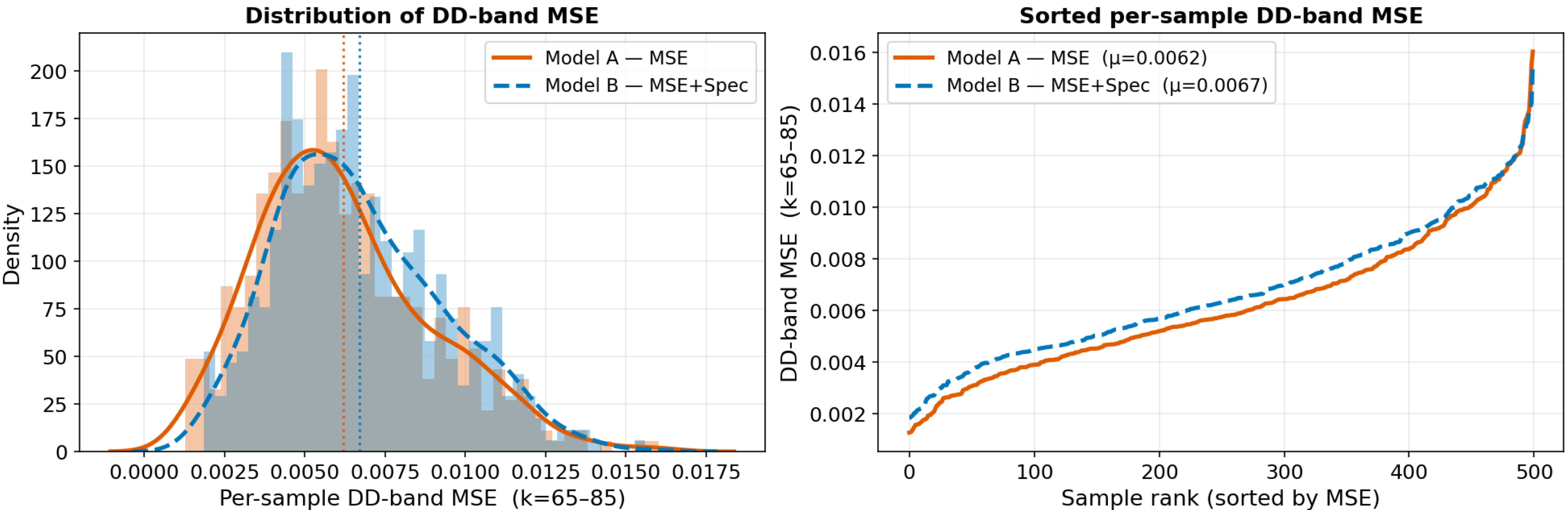}
\caption{Ensemble DD-band pointwise MSE distribution and sorted
per-sample curves. Model~B's slightly larger MSE is systematic
across the ensemble, not driven by outliers; the support--amplitude
decomposition in \cref{fig:dd_decomp} explains why.}
\label{fig:dd_mse}
\end{figure*}

\cref{fig:vae_qualitative} compares a representative held-out field
with its Model~A and Model~B reconstructions and the corresponding
DD-band band-pass fields. \cref{fig:dd_mse} shows the ensemble
distribution of DD-band pointwise MSE underlying the
suppression-vs-recovery analysis of \cref{sec:dd_decomp}.

\end{document}